\documentclass{article}

\usepackage[numbers]{natbib}
\usepackage[preprint]{neurips_2022}

\usepackage[dvipsnames]{xcolor}         
\definecolor{linkColor}{rgb}{0.2,0.4,0.6}
\usepackage[utf8]{inputenc} 
\usepackage[T1]{fontenc}    
\usepackage[colorlinks=true,linkcolor=linkColor,citecolor=linkColor,filecolor=linkColor,urlcolor=linkColor]{hyperref}       
\usepackage{url}            
\usepackage{booktabs}       
\usepackage{amsfonts}       
\usepackage{nicefrac}       
\usepackage{graphicx}
\usepackage{wrapfig}
\usepackage{makecell}
\usepackage{bbm}
\usepackage{float}
\usepackage{subcaption}
\usepackage{soul}
\usepackage{fontawesome5}
\usepackage{pifont}
\usepackage{epstopdf}
\usepackage{cuted}
\usepackage{multirow}
\usepackage{mathtools}
\usepackage{enumitem}
\usepackage{adjustbox}
\usepackage{setspace}
\usepackage{ulem}
\usepackage{capt-of}
\usepackage{algorithm}
\usepackage{algpseudocode}
\usepackage{algorithmicx}
\usepackage{wasysym}
\usepackage[most]{tcolorbox}
\usepackage{amsmath}
\usepackage{mdframed}

\definecolor{gray}{rgb}{0.5,0.5,0.5}
\definecolor{gray94}{gray}{.92}
\definecolor{gray90}{gray}{.90}
\definecolor{gray85}{gray}{.85}
\usepackage{colortbl}
\usepackage{pifont}
\usepackage{amssymb}
\usepackage{ulem} 
\usepackage{adjustbox}
\usepackage{graphicx}
\usepackage{wrapfig}
\usepackage[utf8]{inputenc} 
\usepackage{makecell}
\usepackage{bbm}
\usepackage{float}
\usepackage{epstopdf}
\usepackage{cuted}
\usepackage{multirow}
\usepackage{mathtools}
\usepackage{enumitem}
\usepackage{adjustbox}
\usepackage{setspace}
\usepackage{ulem}
\usepackage{capt-of}
\usepackage[dvipsnames]{xcolor}
\usepackage{algorithm}
\usepackage{algpseudocode}
\usepackage{algorithmicx}
\usepackage{hyperref}
\usepackage{url}
\usepackage{booktabs}
\usepackage[most]{tcolorbox}
\usepackage{amssymb}
\usepackage{arydshln}
\usepackage{graphicx}
\usepackage{listings}
\newcommand\ours{{\texttt{BitDistill}}}

\newcommand\oursfull{{\texttt{BitNet Distillation}}}
\newcommand{\cmark}{{\color{blue}\ding{51}}}%
\newcommand{\xmark}{{\color{red}\ding{55}}}%
\usepackage[table]{xcolor}
\definecolor{darkblue}{rgb}{0, 0, 0.6}

\title{BitNet Distillation}
\author{
Xun Wu~~~Shaohan Huang~~~Wenhui Wang~~~ Ting Song~~~Li Dong~~~Yan Xia~~~Furu Wei\footnotemark[2]
\\\\
Microsoft Research \\
{\href{https://aka.ms/GeneralAI}{https://aka.ms/GeneralAI}}
\vspace{-0.4cm}
\\}

\begin{document}
\maketitle
\begin{abstract}
%
In this paper, we present \oursfull{}~(\ours{}), a lightweight pipeline that fine-tunes off-the-shelf full-precision LLMs (e.g., Qwen) into 1.58-bit precision (i.e., ternary weights \{-1, 0, 1\}) for specific downstream tasks, achieving strong task-specific performance with minimal computational cost. Specifically, \ours{} incorporates three key techniques: the SubLN module, as introduced in BitNet~\cite{wang2023bitnet1bit}; multi-head attention distillation, based on MiniLM~\cite{wang2020minilmv2}; and continual pre-training, which serves as a crucial warm-up step to mitigate the \textbf{scalability issue of the performance gap} between finetuned full-precision and 1.58-bit LLMs on specific tasks.
Experimental results show that \ours{} achieves \textbf{performance comparable to the full-precision counterpart models} across model size, while enabling up to $10\times$ memory savings and $2.65\times$ faster inference on CPUs.
Code is available at \href{https://github.com/microsoft/BitNet}{\texttt{github.com/microsoft/BitNet}}.
\end{abstract}

\begin{figure*}[h]
\includegraphics[width=1.0\linewidth]{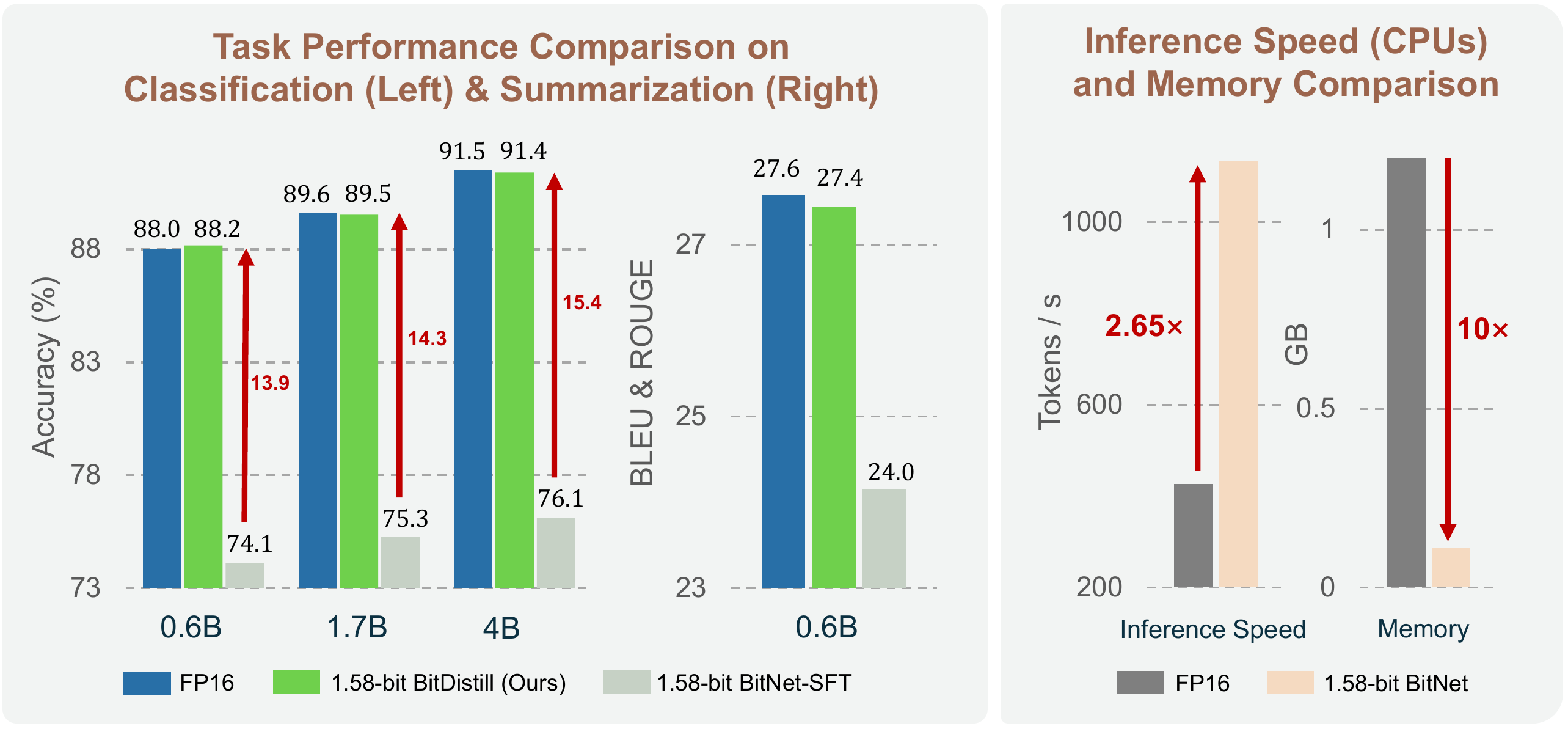}
\caption{\textbf{Performance on downstream tasks across model size, with inference speed and memory efficiency comparison}. We observed that directly finetuning full-precision LLMs into 1.58-bit LLMs (denoted as 1.58-bit BitNet-SFT) leads to a notable performance gap compared to the FP16 baseline, and this gap remains or even widens as the model size increases. In contrast, \ours{} preserves scalability, resulting in performance comparable to full-precision counterparts across all model size, while reducing $10\times$ memory usage and $2.65\times$ faster inference on CPUs.}
\label{fig:intro}
\end{figure*}

\renewcommand{\thefootnote}{\fnsymbol{footnote}}
\footnotetext[2]{Corresponding author.}
\clearpage


\section{Introduction}
\label{sec: intro}
%
Large Language Models (LLMs)~\citep{achiam2023gpt,guo2025deepseekr1} have become indispensable not only in advancing general natural language processing~\citep{yang2025qwen3}, but more importantly in powering a wide range of downstream applications, such as recommendation~\citep{wu2024survey,hou2024large,ren2024representation}, classification~\citep{kostina2025large,sun2023text}, and retrieval~\citep{zhao2024optimizing,borgeaud2022improving}.
Despite their broad applicability, deploying LLMs in downstream applications remains highly challenging. The rapid escalation in model size further amplifies these challenges, especially on resource-constrained devices (e.g., smartphones), where both memory consumption and computational overhead become prohibitive.

To address these challenges, recent efforts on extreme low-bit LLMs, such as the 1.58-bit (i.e., ternary values \{\texttt{-1}, \texttt{0}, \texttt{1}\}) BitNet~\citep{ma2024bitnet158,ma2025bitnet2b4t,wang2023bitnet1bit}, aim to dramatically reduce memory footprint and accelerate inference, offering a promising avenue for efficient deployment in downstream applications.
However, achieving competitive accuracy on downstream applications with 1.58-bit BitNet generally requires pretraining from scratch on large-scale corpora~\citep{team2025minicpm4,ma2025bitnet2b4t} first, resulting in substantial computational and energy overhead.
Furthermore, as illustrated in Figure~\ref{fig:intro}, directly applying quantization-aware training (QAT)~\citep{du2024bitdistiller,chen2024efficientqat} to existing full-precision LLMs at 1.58-bit for specific downstream tasks is often unstable, fails to fully preserve the performance of their full-precision counterparts, and exhibit an poor scalability: as model size increases from 0.6B to 4B, the performance gap relative to the full-precision baseline grows from 13.9 to 15.3. This highlights the pressing need for more effective QAT methods specifically designed for 1.58-bit BitNet.

In this work, we focus on \textbf{fine-tuning existing LLMs to 1.58-bit for specific downstream tasks, while achieving performance comparable to their full-precision counterparts}. To this end, we propose \oursfull{}~(\ours{}), a scaling-friendly QAT framework designed to bridge the gap between extreme 1.58-bit quantization and practical deployment. \ours{} comprises three stages: (i) modeling refinement with SubLN module~\citep{wang2023bitnet1bit} for stable optimization, (ii) continued pre-training to mitigate scale-related performance gaps, and (iii) MiniLM-based~\citep{wang2020minilm,wang2020minilmv2} multi-head attention distillation to recover full-precision accuracy.

Through extensive evaluations across four benchmarks and diverse model scales, we demonstrate that \ours{} scales effectively, achieving downstream task performance on par with full-precision baselines. At the same time, as shown in Figure~\ref{fig:intro}, it reduces $10\times$ memory savings and $2.65\times$ faster inference on CPUs, offering significant improvements in latency, throughput, memory efficiency, and energy consumption, which makes it particularly well-suited for deployment on resource-constrained hardware.


Specifically, this work makes the following contributions:
\begin{enumerate}[leftmargin=1.2em,itemsep=0pt,parsep=0.2em,topsep=0.0em,partopsep=0.0em]
\item To the best of our knowledge, we are the first to investigate fine-tuning pre-trained full-precision LLMs into 1.58-bit BitNet for specific downstream tasks, and we identify key challenges including: performance degradation, poor scalability, and training instability.
\item To address these challenges, we propose a tailored distillation framework named \ours{}, which comprises three key techniques: the SubLN module, as introduced in BitNet~\cite{wang2023bitnet1bit}; multi-head attention distillation, based on MiniLM~\cite{wang2020minilmv2}; and continual pre-training, which serves as a crucial warm-up step to mitigate the scalability issue of the performance gap between finetuned full-precision and 1.58-bit LLMs on specific tasks.
\item Extensive experiments across multiple benchmarks and model scales show that \ours{} enables 1.58-bit quantized LLMs to achieve downstream performance comparable to their full-precision counterparts, while enabling up to $10\times$ memory savings and $2.65\times$ faster inference on CPUs.
\end{enumerate}

\section{Preliminaries}

\noindent\textbf{1.58-bit Quantization}. Following~\citep{ma2024bitnet158}, we adopt per-tensor quantization using the \texttt{absmean} function to map the weights of existing LLMs into ternary values, i.e., \{\texttt{-1}, \texttt{0}, \texttt{1}\}:
\begin{align}
    &\text{Q}_{w}(\mathbf{W}) = \Delta~\text{RoundClip}(\frac{\mathbf{W}_{\text{FP16}}}{\Delta+\epsilon}, -1, 1),\, \\
    \quad\text{where} ~\Delta =~& \text{mean}(|\mathbf{W}|), \quad\text{RoundClip}(\mathbf{Y}, a, b) = \min\left(\max\left(\lfloor \mathbf{Y}\rceil, a\right), b\right),
    \label{eq: weight}
\end{align}
The notation $\lfloor \mathbf{\cdot}\rceil$ means the nearest rounding operation. For LLM inputs, we employ 8-bit activation quantization. Specifically, we use per-token \texttt{absmax} and \texttt{absmean} functions to quantize the activations, which can be formulated as:
\begin{equation}
    \text{Q}_{\text{INT8}}(\mathbf{X}) = \frac{\gamma}{127}~\text{RoundClip}(\frac{127}{\gamma+\epsilon}\mathbf{X}_{\text{FP16}}, -128, 127),\,\gamma = \max(|\mathbf{X}_{\text{FP16}}|)
    \label{eq: activation}
\end{equation}
\noindent\textbf{Gradient Approximation}. Due to the presence of non-differentiable operations in Eq.~\ref{eq: weight} and Eq.~\ref{eq: activation} (e.g., \text{RoundClip}), the gradient cannot be propagated through the entire model during backpropagation. Following~\citep{ma2024bitnet158,ma2025bitnet2b4t,wang2023bitnet1bit}, we employ the Straight-Through Estimator (STE)~\citep{bengio2013ste} to approximate gradients for 1.58-bit quantized LLMs.
\section{\ours{}: Finetuning LLMs into 1.58-bit BitNet For Downstream Tasks}
\label{sec: BDF}

In this work, we address the challenge of deploying LLMs on resource-constrained devices for specific downstream tasks. We focus on efficiently compressing existing pre-trained LLMs to 1.58-bit BitNet with minimal performance degradation and training cost. Our proposed~\oursfull{} (\ours{}) incorporates three key stages: (1) \textbf{Modeling refinement} with SubLN~\citep{wang2023bitnet1bit} for stable optimization (detailed in~\S~\ref{sec: modeling}), (2) \textbf{Continue pre-training} as a crucial warm-up step to mitigate the performance gap that does not scale well between fine-tuned full-precision models and 1.58-bit BitNet (see in~\S~\ref{sec: training}), and (3) \textbf{Distillation-based fine-tuning}, which leverages both logits distillation and multi-head attention distillation to recover full-precision performance (see~\S\ref{sec: distill}).

\subsection{Stage-1: Modeling Refinement}
\label{sec: modeling}
Unlike full-precision models, where the variance of hidden states is typically preserved within a stable range under standard initialization schemes, low-bit quantized models such as 1.58-bit LLMs often suffer from excessively large activation variance, which results in optimization instability and degraded convergence~\citep{ma2024bitnet158,wang2023bitnet1bit}. 

To alleviate this issue, following the design principles of prior 1.58-bit BitNet~\citep{ma2024bitnet158,ma2025bitnet2b4t}, we introduce additional normalization layers named SubLN at carefully chosen positions inside each transformer block. Specifically, instead of only applying pre-normalization at the block input, we further insert SubLN right before the output projection of the Multi-Head Self-Attention (MHSA) module as well as before the output projection of the Feed-Forward Network (FFN). Concretely, taking Qwen3~\citep{yang2025qwen3} as a reference architecture, the computations at the $l$-th transformer layer are modified as:
\begin{gather}
\mathbf{Y}_l 
= \mathbf{X}_l + 
        {\color[rgb]{0,0,1.0}\textbf{SubLN}}\bigl(
            \text{Concat}(\text{heads})
        \bigr)\mathbf{W}^{\text{MHSA}}_{\text{out}}, 
\\
\mathbf{X}_{l+1} 
= \mathbf{Y}_l  +  {\color[rgb]{0,0,1.0}\textbf{SubLN}}\bigl(
            (\,\mathbf{Y}_l \mathbf{W}^{\text{FFN}}_{\text{up}}\,) \odot \sigma(\mathbf{Y}_l \mathbf{W}^{\text{FFN}}_{\text{gate}})
        \bigr) \mathbf{W}^{\text{FFN}}_{\text{down}},
\end{gather}
where
\begin{gather}
\text{heads} = \Bigl\{
    \text{Softmax}\!\left(\frac{\mathbf{Q}_i\mathbf{K}_i^\top}{\sqrt{d}}\right)\mathbf{V}_i
   \;\Big|\; \mathbf{Q}_i = \mathbf{X}\mathbf{W}^{\text{MHSA}}_{Q,i},\, \mathbf{K}_i = \mathbf{X}\mathbf{W}^{\text{MHSA}}_{K,i},\, \mathbf{V}_i = \mathbf{X}\mathbf{W}^{\text{MHSA}}_{V,i}
\Bigr\},
\end{gather}

where the outer SubLN in each equation corresponds to the newly inserted normalization before the respective output projection. This design ensures that the hidden representations entering quantized projection layers are variance-stabilized, preventing the explosion of activation scale and thereby improving both training stability and task performance.

\subsection{Stage-2: Continue Pre-Training}
\label{sec: training}
As shown in Figure~\ref{fig:intro}, directly fine-tuning 1.58-bit BitNet modified from existing full-precision LLMs on downstream tasks may yield suboptimal results, as the limited number of training tokens is often insufficient to effectively adapt full-precision weights into the constrained 1.58-bit representation, which leads to exhibit poor scalability: as model size increases, the performance gap relative to the full-precision baseline widens.

To this end, we propose a two-stage training pipeline consisting of a continue training stage, which leverages only a small amount of pretraining corpus to achieve the desired adaptation, followed by fine-tuning on the downstream task.
Specifically, given a small set of corpus $\mathbf{C} = \{\mathbf{c}_1, \cdots, \mathbf{c}_{N}\}$, we finetuning the modeling-modified pre-trained LLMs attained from \S~\ref{sec: modeling} as:
\begin{equation}
\mathcal{L}_{\text{CT}} = - \frac{1}{N} \sum_{i=1}^{N} \sum_{t=1}^{T_i} \log P_{\theta}\!\left(\mathbf{c}_{i,t} \mid \mathbf{c}_{i,<t}\right).
\end{equation}
Here $P_{\theta}$ denotes the probability distribution parameterized by the model.
A detailed analysis of the effect of continue training, along with an investigation into the underlying mechanisms and supporting hypotheses, can be found in~\S~\ref{sec: analysis}.

\subsection{Stage-3: Distillation-based Fine-tuning}
\label{sec: distill}
To better mitigate the performance degradation introduced by precision reduction, we incorporate two kinds of knowledge distillation technology into the downstream task finetuning phase, where the fine-tuned full-precision LLMs serves as the teacher and its 1.58-bit quantized counterpart acts as the student.

\noindent\textbf{Logits Distillation}. Logits distillation has recently been widely adopted in the QAT phase of quantized models, demonstrating promising effectiveness~\cite{du2024bitdistiller,lee2025unifying,ko2024distillm}. Given data pairs $\{(\mathbf{x}_i, \mathbf{y}_i)\}_{i=1}^{N}$ sampled from downstream datasets, 
the objective of logits distillation is defined as
\begin{gather}
\mathcal{L}_{\text{LD}}
= \frac{1}{N} \sum_{i=1}^{N} 
\mathcal{D}_{\text{KL}}\!\left(
P_{\theta}^{\text{FP16}}\!\left(\mathbf{y}_i \mid \mathbf{x}_i\right)
\;\big\|\;
P_{\theta}^{\text{1.58-bit}}\!\left(\mathbf{y}_i \mid \mathbf{x}_i\right)
\right),
\\
P^{(\cdot)}_{\theta}(\mathbf{y} \mid \mathbf{x}) 
= \frac{\exp\!\left( z_y(\mathbf{x}; \theta) / \tau \right)}
{\sum_{y'} \exp\!\left( z_{y'}(\mathbf{x}; \theta) / \tau \right)}
\label{eq: logits distill}
\end{gather}
Here $z_y(\mathbf{x}; \theta)$ denotes the unnormalized logit produced by the model for $y$ when given the input $\mathbf{x}$. The temperature parameter $\tau$ is introduced to control the softening of the output distributions for both the FP16 and 1.58-bit models. $\mathcal{D}_{\text{KL}}(\cdot\parallel\cdot)$ represents the Kullback–Leibler divergence.

\noindent\textbf{Multi-Head Attention Distillation,}. Since the attention mechanism plays a pivotal role in LLMs and largely determines their overall performance, we further investigate distillation at the attention layer to encourages the 1.58-bit student to capture the fine-grained structural dependencies embedded in the FP16 teacher’s attention patterns.

Following MiniLM series~\citep{wang2020minilmv2,wang2020minilm}, given training samples $\mathbf{x}$ drawn from the downstream dataset, we define the attention-relations distillation loss $\mathcal{L}_{\text{AD}}$ as
\begin{gather}
\mathbf{A}^{(\cdot)} \sim \Phi, \quad
\Phi = \{\mathbf{Q}, \mathbf{K}, \mathbf{V}\}, \\
\mathcal{L}_{\text{AD}} = \frac{1}{|\Upsilon|}\sum_{i=1}^{|\Upsilon|}
\sum_{j=1}^{|\Phi|}{\alpha_{i}\frac{1}{A_{r}|\mathbf{x}|}
\sum_{a=1}^{A_{r}}\sum_{t=1}^{|\mathbf{x}|}
\mathcal{D}_{\text{KL}}(\mathbf{R}^{\text{FP16}}_{i,j,a,t} \parallel \mathbf{R}^{\text{1.58-bit}}_{i,j,a,t})}.
\end{gather}
Here $\Phi$ correspond to the query, key, and value projections within a multi-head attention block, and $\Upsilon$ denotes the set of layers we selected for distillation. $\alpha_{i}$ are coefficients controlling the relative weights of different relational terms. The sequence length is denoted by $|\mathbf{x}|$, $A_{r}$ is the number of attention heads. The relational distribution $\mathbf{R}^{(\cdot)}_{i,j,a,t}$ is derived by applying scaled dot-product attention followed by $\text{Softmax}$ with hidden dimension $d_r$, while $\mathbf{R}^{\text{1.58-bit}}_{i,j,a,t}$ is obtained analogously from the quantized student model using hidden dimension $d_r^{\prime}$, i.e.,
\begin{gather}
\mathbf{R}_{i,j,a,t}^\text{FP16} = \mathrm{Softmax}\!\left(\frac{\mathbf{A}_{i,j,a,t}^{\text{FP16}} {\mathbf{A}_{i,j,a,t}^{\text{FP16}}}^\top}{\sqrt{d_r}}\right), \quad
\mathbf{R}_{i,j,a,t}^\text{1.58-bit} = \mathrm{Softmax}\!\left(\frac{\mathbf{A}_{i,j,a,t}^{\text{1.58-bit}} {\mathbf{A}_{i,j,a,t}^{\text{1.58-bit}}}^\top}{\sqrt{d_r^{\prime}}}\right).
\label{eq: attn distill}
\end{gather}
%

The detailed implement of $\mathcal{L}_{\text{AD}}$ can be found in Algorithm~\ref{alg: L_ad}.
Following MiniLM~\citep{wang2020minilm,wang2020minilmv2}, \textbf{we recommend performing attention distillation at only a single layer (i.e., $|\Upsilon| = 1$) rather than across all layers}, as conferring greater optimization flexibility to the 1.58-bit student BitNet often yields superior downstream performance.

The total loss of the distillation-based finetuning phase $\mathcal{L}$ comprises three terms that aim to minimize the discrepancy between the student and teacher models and improve downstream task performance, scaled by two distillation coefficients, $\lambda$ and $\gamma$, i.e.,
\begin{gather}
\mathcal{L} = \mathcal{L}_{\text{CE}} + \lambda\mathcal{L}_{\text{LD}} + \gamma\mathcal{L}_{\text{AD}},
\\
\text{where} \quad \mathcal{L}_{\text{CE}} = - \frac{1}{N} \sum_{i=1}^{N} \sum_{t=1}^{|\mathbf{y}_i|} \log P_{\theta}\!\left(\mathbf{y}^{t}_{i} \mid \mathbf{x}_{i}\right).
\label{eq: loss sum}
\end{gather}
Here $\mathcal{L}_{\text{CE}}$ denotes the cross-entropy loss on the downstream dataset. $\lambda$ and $\gamma$ control the trade-off between distillation and model fitting.

\begin{algorithm}[t]
\scriptsize
\caption{Pseudo Torch Style Implement of $\mathcal{L}_{\text{AD}}$}
\label{alg: L_ad}
\definecolor{codeblue}{rgb}{0.25,0.5,0.5}
\definecolor{codegreen}{rgb}{0,0.6,0}
\definecolor{codekw}{rgb}{0.85, 0.18, 0.50}
\lstset{
  backgroundcolor=\color{white},
  basicstyle=\fontsize{7.5pt}{7.5pt}\ttfamily\selectfont,
  columns=fullflexible,
  breaklines=true,
  captionpos=b,
  commentstyle=\fontsize{7.5pt}{7.5pt}\color{codegreen},
  keywordstyle=\fontsize{7.5pt}{7.5pt}\color{codekw},
  escapechar={|}, 
}
\begin{lstlisting}[language=python]
def compute_attention_distillation_loss(student_states, teacher_states, distill_layer, split_heads):
  # student_states [3, B, num_heads, seq_len, head_dim]: Q, K, V states from the 1.58-bit model
  # teacher_states [3, B, num_heads, seq_len, head_dim]: Q, K, V states from the FP16 model
  # distill_layer [1]: the index of layers used for distillation
  # split_heads [1]: the number of heads when computing attention relation matrix
  _, B, heads, L, d = student_states.shape
  D = heads * d // split_heads
  # Loop for computing distillation loss across Q, K, V
  for i in range(3):
    s_values, t_values = student_states[i], teacher_states[i]
    s_values = F.normalize(s_values.transpose(1, 2).reshape(B, L, split_head, D).transpose(1, 2), dim=-1)
    t_values = F.normalize(t_values.transpose(1, 2).reshape(B, L, split_head, D).transpose(1, 2), dim=-1)
    # Compute relation martix
    s_relation = torch.matmul(s_values, s_values.transpose(-2, -1))
    t_relation = torch.matmul(t_values, t_values.transpose(-2, -1))
    # Reshape: [B, split_heads, L, L] -> [B*split_heads*L, L]
    s_relation = (s_relation / temperature).reshape(-1, L)
    t_relation = (t_relation / temperature).reshape(-1, L)
    
    s_prob = F.softmax(s_relation, dim=-1).clamp(min=1e-8)
    t_prob = F.softmax(t_relation, dim=-1).clamp(min=1e-8)

    distill_loss += F.kl_div(torch.log(s_prob), t_prob, reduction="batchmean", log_target=False)
  return distill_loss
\end{lstlisting}
\end{algorithm}

\section{Experiments}

\subsection{Experimental Setup}

\noindent\textbf{Datasets}. We evaluate the effectiveness of our proposed method, \ours{}, on two representative tasks: \textbf{text classification} and \textbf{text summarization}. For classification, we adopt three widely used datasets from the General Language Understanding Evaluation (GLUE) benchmark~\citep{wang2018glue}\footnote{\href{https://gluebenchmark.com/}{https://gluebenchmark.com/}}: the Multi-Genre Natural Language Inference Corpus (MNLI)~\citep{mnli}, the Question-answering Natural Language Inference dataset (QNLI)~\citep{qnli}, and the Stanford Sentiment Treebank (SST-2)~\citep{sst2}. These datasets are employed for both training and evaluation to comprehensively assess the effectiveness of our approach. For summarization, we use the CNN/DailyMail dataset (CNNDM)~\citep{cnndm}\footnote{\href{https://huggingface.co/datasets/abisee/cnn_dailymail}{https://huggingface.co/datasets/abisee/cnndailymail}} as both the training and evaluation corpus.

\noindent\textbf{Baselines for Comparison}. Since our objective is to fine-tune pre-trained full-precision LLMs into 1.58-bit BitNet models for specific downstream tasks, we compare the performance of our 1.58-bit models (denoted as \textbf{\ours{}}) with that of FP16 models fine-tuned directly on the corresponding downstream tasks (named \textbf{FP16-SFT}). In addition, we also report the results of directly converting full-precision LLMs into 1.58-bit BitNet models and fine-tuning them on downstream tasks (denoted as \textbf{BitNet-SFT}).

\noindent\textbf{Training Settings}. We fine-tune the Qwen3 series~\citep{yang2025qwen3} as our base models, covering 0.6B, 1.7B, and 4B parameter scales. In addition, we investigate the impact of different base model types by conducting experiments with alternative backbones such as Gemma~\citep{team2025gemma} and Qwen2.5~\citep{qwen2025qwen25}. For all baseline methods and our approach, we adopt a greedy search strategy to select the optimal learning rate and training epochs. This procedure mitigates overfitting while ensuring both strong downstream performance and fair comparisons across methods. 
We fix the maximum training sequence length to 512 tokens and the batch size to 32. All models are trained on servers equipped with 8$\times$AMD Mi300X GPUs.

Specifically, we set the temperature for logits distillation (Eq.~\ref{eq: logits distill}) to 5.0. For the classification task, we use $\lambda = 10$ and $\gamma = 1\text{e}5$ in Eq.~\ref{eq: loss sum}, while for the summarization task, we set $\lambda = 1$ and $\gamma = 1\text{e}3$. We set $\alpha_i = 1.0$ for all experiments. During the continue pre-training phase described in~\S\ref{sec: training}, we further train our models using only 10B tokens sampled from the FALCON corpus~\citep{penedo2023refinedwebdatasetfalconllm}. Compared with the cost of pre-training a 1.58-bit BitNet from scratch (approximately 4T tokens)~\citep{ma2025bitnet2b4t}, this additional cost is virtually negligible.

\noindent\textbf{Evaluation Settings}. For both classification and summarization task, we fix the sampling parameters by setting top-$p$ to 1.0 and the temperature to 0. Classification performance is evaluated using accuracy. For the summarization task, we set the maximum generation length to 4096 tokens. Summarization quality is assessed using BLEU~\citep{papineni2002bleu} and ROUGE-1, ROUGE-2, ROUGE-L and ROUGE-SUM~\citep{lin2004rouge}. For model runtime efficiency, we report the token throughput (tokens per second) on CPU with 16 threads.

\subsection{Main Results}
\begin{table*}[t!]
\centering
\caption{\textbf{Results on text classification tasks.} All models are initialized from the Qwen3 series~\citep{qwen2025qwen25}. The top scores for each metric and dataset are highlighted in bold. The 1.58-bit \ours{} models achieve performance comparable to the FP16 baseline while providing 2× faster inference and 10× memory reduction across all datasets. $^*$ denotes the FP16 teacher used in \ours{}.}
\resizebox{\textwidth}{!}{
\begin{tabular}{lccccccccccc}
\toprule[1.2pt]
\multirow{2}{*}{\textbf{Method}} & \multicolumn{3}{c}{\textbf{MNLI}}    & \multicolumn{3}{c}{\textbf{QNLI}}         &  \multicolumn{3}{c}{\textbf{SST2}}  & \multicolumn{1}{c}{\textbf{Speed}}  & \multicolumn{1}{c}{\textbf{Memory}} \\
\cmidrule(lr){2-4} \cmidrule(lr){5-7} \cmidrule(lr){8-10} \cmidrule(lr){11-11} \cmidrule(lr){12-12}
& 0.6B & 1.7B & 4B &  0.6B & 1.7B & 4B & 0.6B & 1.7B & 4B & (tokens / s) & (G) \\
\midrule
FP16-SFT $^*$ & 88.01 & 89.61 & 91.48 & 93.72 & 95.00 & 96.02 & 94.21 & 95.43 & 96.57 & 427 & 1.20\\
\cdashline{1-12}[2pt/2pt]
\rule{0pt}{10pt}%
BitNet-SFT & 74.09 & 75.27 & 76.11 & 78.32 & 79.54 & 79.97 & 79.92 & 81.37 & 82.07 & \bf 1,135 & \bf 0.11\\
\ours{} (Ours) & \bf 88.17 & \bf 89.53 & \bf 91.40 & \bf 93.66 & \bf 94.82 & \bf 95.93 & \bf 94.30 & \bf 95.26 & \bf 96.47 & \bf 1,135  & \bf 0.11\\
\bottomrule[1.2pt]
\end{tabular}
}
\label{tab:main-benchmark-tc}
\end{table*}
\begin{table*}[t!]
\centering
\renewcommand\tabcolsep{3.0pt}
\caption{\textbf{Results on text summarization tasks (CNNDM dataset).} All models are initialized from the Qwen3 series~\citep{qwen2025qwen25}. The top scores for each metric and dataset are highlighted in bold. The 1.58-bit \ours{} models achieve performance comparable to the FP16 baseline while providing 2× faster inference and 10× memory reduction across all datasets. $^*$ denotes the FP16 teacher used in \ours{}.}
\resizebox{\textwidth}{!}{
\begin{tabular}{lcccccccc}
\toprule[1.2pt]
\bf Method & BLEU & ROUGE-1 & ROUGE-2 & ROUGE-L & ROUGE-SUM & AVG & \multicolumn{1}{c}{\textbf{Speed}~(tokens / s)}  & \multicolumn{1}{c}{\textbf{Memory} (G)} \\
\midrule
FP16-SFT $^*$   & 13.98 & 40.62 & 17.77 & 27.72 & 37.80 & 27.58 & 427 & 1.20\\
\cdashline{1-9}[2pt/2pt]
\rule{0pt}{10pt}%
BitNet-SFT & 11.47 & 37.10 & 13.97 & 24.84 & 33.37 & 24.15 & \bf 1,135  & \bf 0.11 \\
\ours{} (Ours) & \bf 14.41 & \bf 40.21 & \bf 17.47 & \bf 27.49 & \bf 37.63 & \bf 27.44 & \bf 1,135  & \bf 0.11\\
\bottomrule[1.2pt]
\end{tabular}
}
\label{tab:main-benchmark-ts}
\end{table*}
\noindent\textbf{Overall Performance.} The overall evaluation results on the benchmark datasets are reported in Table~\ref{tab:main-benchmark-tc} and Table~\ref{tab:main-benchmark-ts}. Across different model sizes and tasks, the proposed 1.58-bit BitNet models trained with our distillation framework (\ours{}) demonstrate accuracy that is largely comparable to their full-precision counterparts, with only marginal differences observed in most cases. At the same time, the 1.58-bit models deliver substantial gains in system efficiency, including up to a $2\times$ inference speedup on CPUs and nearly an order-of-magnitude reduction in memory footprint. These improvements underline the practical utility of our approach for scenarios where computational resources are constrained, while also showing that aggressive quantization can be made viable with carefully designed distillation strategies.

\noindent\textbf{Robustness to Different Pretrained Models.} To further examine the generality of our framework, we extend the evaluation by replacing the Qwen3 series with alternative base models such as Qwen2.5~\citep{qwen2025qwen25}\footnote{\href{https://huggingface.co/Qwen/Qwen2.5-0.5B}{https://huggingface.co/Qwen/Qwen2.5-0.5B}} and Gemma~\citep{team2025gemma}\footnote{\href{https://huggingface.co/google/gemma-3-1b-pt}{https://huggingface.co/google/gemma-3-1b-pt}}. The results, summarized in Table~\ref{tab:main-benchmark-base-model-left}, indicate that \ours{} consistently yields downstream performance close to that of full-precision fine-tuning across all examined architectures. While minor performance fluctuations are observed between base models, the trend remains stable, suggesting that our method is not tailored to a specific pretraining family but can be applied more broadly. This robustness enhances the potential applicability of our approach in diverse deployment environments, where the choice of pretrained backbone may vary depending on availability and task requirements.
\subsection{Ablation Study}
\begin{table*}[t!]
\centering
\begin{minipage}{0.48\linewidth}  
\centering
\caption{Results on the text classification task (MNLI dataset) with different base model initializations. $^*$ denotes the FP16 teacher used in \ours{}.}
\resizebox{\linewidth}{!}{
\begin{tabular}{lcc}
\toprule[1.2pt]
Method & Gemma3-1B & Qwen2.5-0.5B \\
\midrule
FP16-SFT $^*$  & 89.77 & 79.91 \\
\cdashline{1-3}[2pt/2pt]
\rule{0pt}{10pt}%
BitNet-SFT & 78.02 & 60.80  \\
\ours{} & \textbf{89.61} & \textbf{79.98}  \\
\bottomrule[1.2pt]
\end{tabular}
}
\label{tab:main-benchmark-base-model-left}
\end{minipage}
\hfill
\begin{minipage}{0.48\linewidth}  
\centering
\caption{Results on the text classification task with different quantization techniques. B, G, A indicates Block Quant, GPTQ and AWQ, respectively.}
\resizebox{\linewidth}{!}{
\renewcommand\tabcolsep{8.0pt}
\begin{tabular}{lcc}
\toprule[1.2pt]
Method & MNLI & QNLI \\
\midrule
\ours{} & 88.17 & 93.66  \\
\ours{}-B~\citep{dettmers20218} & 88.23 & 93.74 \\
\ours{}-G~\citep{frantar2022gptq} & 88.05 & 93.63 \\
\ours{}-A~\citep{lin2024awq} & \bf 88.25 & \bf 93.70\\
\bottomrule[1.2pt]
\end{tabular}
}
\label{tab:main-benchmark-quant}
\end{minipage}
\end{table*}

\noindent\textbf{Effect of each individual stages in}~\ours{}.  
As outlined in~\S\ref{sec: BDF}, the \ours{} framework consists of three stages. To understand the contribution of each component, we conduct an ablation study by removing one stage at a time and re-training the model. The results, reported in Table~\ref{tab:main-benchmark-stage}, show that excluding any stage consistently leads to a non-trivial drop in downstream performance. This suggests that each stage plays a complementary role, and that the full pipeline is necessary to obtain the best trade-off between efficiency and accuracy.

\noindent\textbf{Effect of different distillation techniques in Stage-3}~\S\ref{sec: distill}.  
In the final stage of our framework, we introduce two complementary distillation techniques to better optimize 1.58-bit BitNet models for downstream tasks. To disentangle their respective effects, we compare using each technique individually against the joint application of both. As shown in Table~\ref{tab:main-benchmark-distill}, while each technique alone provides partial improvements, the combination leads to the most consistent performance across benchmarks. This observation provides evidence that the two techniques address different aspects of the optimization challenge, and their synergy is particularly beneficial under extreme quantization.

\noindent\textbf{Compatibility with different quantization techniques}.  
We further examine the compatibility of \ours{} with existing post-training and weight-quantization approaches. In particular, we consider Block-Quant~\citep{dettmers20218}, GPTQ~\citep{frantar2022gptq}, AWQ~\citep{lin2024awq}, as well as the simple min–max quantization scheme in Eq.~\ref{eq: weight}. To this end, we integrate \ours{} with each quantization method and evaluate the resulting 1.58-bit models. The results are summarized in Table~\ref{tab:main-benchmark-quant} and lead to two main observations: (1) regardless of the underlying quantization method, models benefit consistently from the proposed framework and generally match the full-precision baseline, and (2) more sophisticated quantization strategies (e.g., GPTQ, AWQ) provide additional gains on top of our distillation pipeline. These findings suggest that \ours{} is complementary to different quantization algorithms, offering a unified procedure that can stably enhance low-bit models across a diverse range of quantization settings.

\subsection{Analysis}
\label{sec: analysis}



\begin{figure*}[t]
\centering
\includegraphics[width=\linewidth]{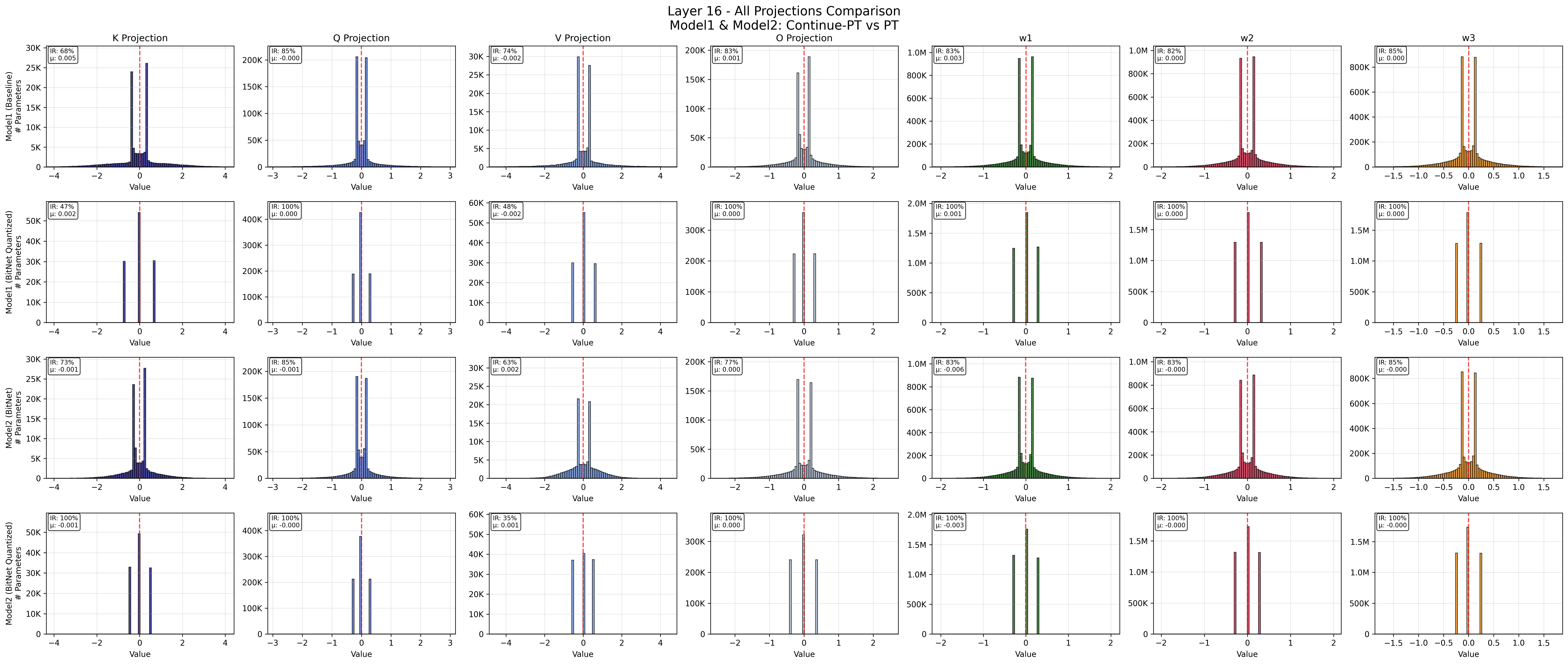}
\\
\makebox[0.135\textwidth]{\footnotesize Attention-K}
\makebox[0.135\textwidth]{\footnotesize Attention-Q}
\makebox[0.135\textwidth]{\footnotesize Attention-V}
\makebox[0.135\textwidth]{\footnotesize Attention-O}
\makebox[0.135\textwidth]{\footnotesize FFN-Gate}
\makebox[0.135\textwidth]{\footnotesize FFN-Up}
\makebox[0.135\textwidth]{\footnotesize FFN-Down}
\caption{\textbf{Visualization of model weights}. The top two rows show the quantized weights of BitNet trained from scratch along with their corresponding FP16 distributions. The bottom two rows show the quantized weights of BitNet after loading weights from LLMs and performing continued training (stage-2 in~\S~\ref{sec: training}), together with their corresponding FP16 distributions.} 
\label{fig: model weight}
\end{figure*}

\begin{figure}[t!] \centering
\includegraphics[width=0.32\textwidth]{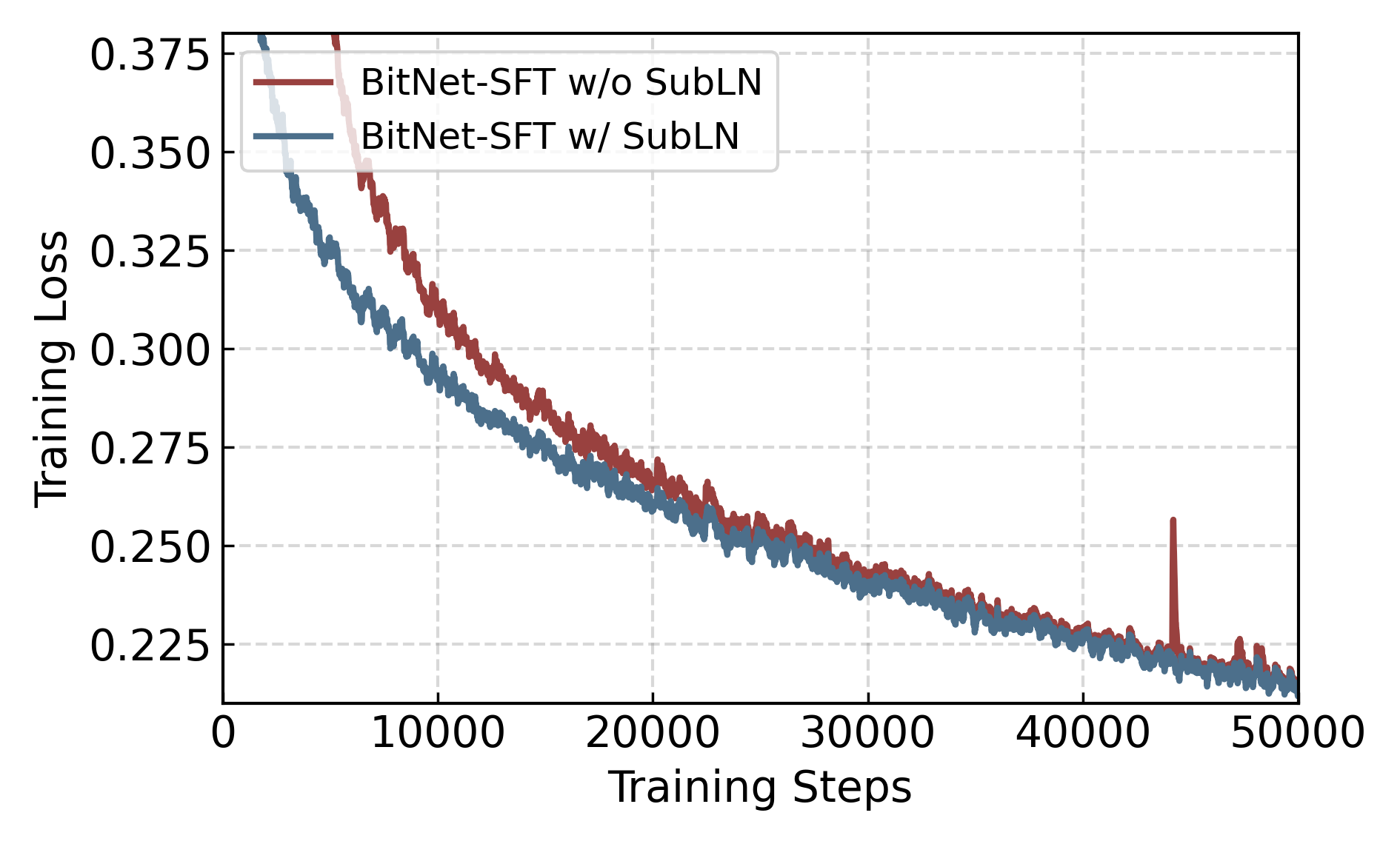}
\includegraphics[width=0.32\textwidth]{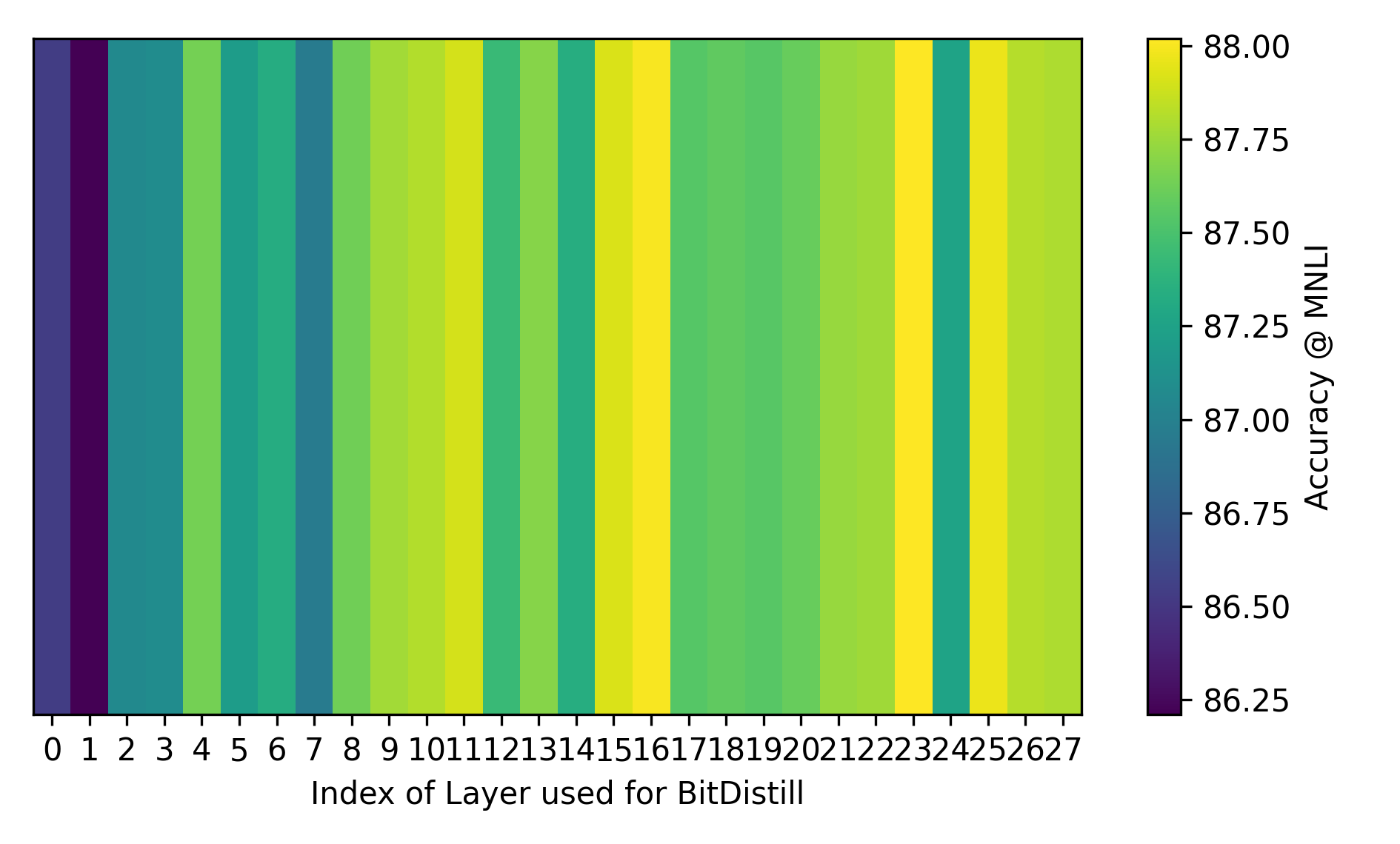}
\includegraphics[width=0.32\textwidth]{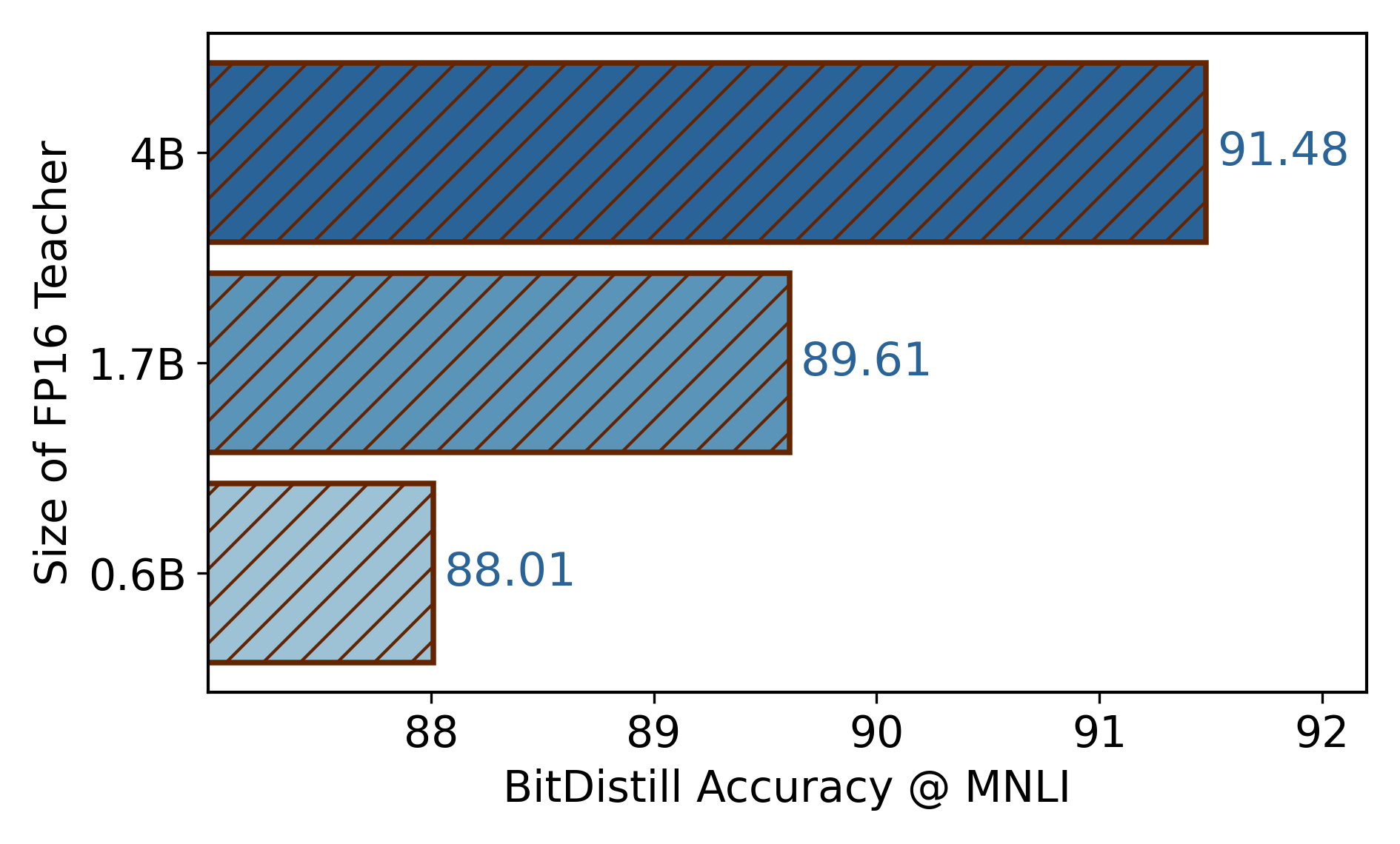}
\\
\makebox[0.32\textwidth]{\small (a)}
\makebox[0.32\textwidth]{\small (b)}
\makebox[0.32\textwidth]{\small (c)}
\\
\caption{\textbf{Analysis of SubLN, layer selection for Eq.~\ref{eq: attn distill} and teacher selection over training steps}. (a) Fine-tuning existing LLMs into 1.58-bit BitNet with SubLN yields better performance and faster convergence. (b) Comparison of MNLI accuracies obtained by distilling different layers on Qwen3-0.6B. (c) Comparison of MNLI accuracies obtained by distilling Qwen3-0.6B with different size of FP16 teachers.} 
\label{fig: analysis}
\end{figure}

\noindent\textbf{Effect of SubLN used in Stage-1 \S~\ref{sec: modeling}}. To validate the effect of SubLN, we quantize existing LLMs into 1.58-bit BitNet and fine-tune them on FALCON corpus, comparing the performance with (denoted as BitNet-SFT w/ SubLN) and without the insertion of SubLN (denoted as BitNet-SFT w/o SubLN). Specifically, as shown by the training loss curve in Figure~\ref{fig: analysis} (a), we find that the modeling refinement detailed in Stage-1 \S~\ref{sec: modeling}, which modifies the LLMs' architecture by inserting SubLN layers at specific positions, effectively stabilizes the optimization of the 1.58-bit BitNet and leads to improved performance.

\noindent\textbf{Why continue-training mitigates the scalability issue}. As stated in \S~\ref{sec: intro}, a critical challenge in applying 1.58-bit BitNet to downstream tasks is the poor scalability, i.e., as model size increases, the performance gap between the 1.58-bit BitNet and its FP16 counterpart becomes increasingly pronounced. Our experiments reveal that a small amount of continue-training can effectively alleviate this issue, and here we investigate the underlying reasons. 

In Figure~\ref{fig: model weight}, we visualize the model weights of 1.58-bit BitNet before and after continue-training, and compare them with those of a BitNet trained from scratch. We find that after continue-training, the weight distribution which initially exhibited an approximately Gaussian shape, becomes more similar to that of a BitNet trained from scratch. This observation supports our hypothesis in \S~\ref{sec: training}: continue-training enables BitNet models to rapidly adapt to the feature space that is better suited for 1.58-bit optimization, thereby preventing convergence to suboptimal local minima and ultimately leading to improved downstream performance.

Furthermore, we investigate why the BitNet-like weight distribution observed in Figure~\ref{fig: model weight} facilitates improved performance on downstream tasks. In particular, the unique distribution concentrates more weights near the transition boundaries between 0 and -1 as well as between 0 and 1. Such placements allow the quantized values to shift more frequently with small gradient steps, thereby enhancing the 1.58-bit BitNet’s ability to fit downstream data and reducing the risk of being trapped in suboptimal local minima.

\noindent\textbf{Distillation layer selection strategy in Stage-3~\S~\ref{sec: distill}}. As discussed in \S~\ref{sec: distill}, we hypothesize that performing attention relation distillation on a single layer provides the 1.58-bit BitNet with greater optimization flexibility compared to distilling across all layers, thereby yielding better performance. To examine this, we explore strategies for selecting the distillation layer. Figure~\ref{fig: analysis} (b) visualizes the MNLI classification results of Qwen3-0.6B when applying distillation to different layers without continue pre-training. Our findings can be summarized as follows: (1) distilling from a single layer achieves superior performance compared to using all layers, supporting our hypothesis; (2) the results vary significantly depending on which single layer is chosen, indicating that an appropriate layer selection strategy is crucial; and (3) layers located in the later stages of the model tend to deliver better distillation performance. 

\noindent\textbf{Better teacher lead to better results.} We investigate whether our proposed \ours{} can leverage a higher-quality FP16 teacher to provide greater downstream task gains for the 1.58-bit BitNet. To this end, we use Qwen3-1.7B and Qwen3-4B FP16 models as teachers in the distillation process for the Qwen3-0.6B 1.58-bit BitNet. The results are visualized in Figure~\ref{fig: analysis} (c). We find that our algorithm can effectively extract larger gains from a higher-quality teacher, even surpassing FP16 models of the same size. This provides a performance guarantee for deploying BitNet models tailored to specific tasks.

\begin{table*}[t!]
\centering
\begin{minipage}{0.67\textwidth}
\centering
\caption{\textbf{Effect of different stages in}~\ours{}. 
Here Qwen3-0.6B is used as base model. 
M.D., C.T., and D.T. denote modeling refinement \S~\ref{sec: modeling}, continue pre-training \S~\ref{sec: training}, and distillation-based finetuning \S~\ref{sec: distill}, respectively.}
\resizebox{\linewidth}{!}{
\begin{tabular}{ccccccccc}
\toprule[1.2pt]
Stage-1 & Stage-2 & Stage-3 & \bf MNLI & \multicolumn{4}{c}{\textbf{CNNDM}} \\
\cmidrule(lr){4-4} \cmidrule(lr){5-8}
M.D. & C.T. & D.F. & ACC & BLEU & ROUGE-1 & ROUGE-2 & ROUGE-L\\
\midrule
\xmark & \xmark & \xmark & 74.09 & 11.47 & 37.10 & 13.97 & 24.84 \\
\cmark & \xmark & \xmark & 76.30 & 11.69 & 37.81 & 14.13 & 25.11 \\
\cmark & \cmark & \xmark & 86.73 & 13.96 & 39.75 & 16.47 & 26.96 \\
\cmark & \xmark & \cmark &  88.04 & 13.70 & 39.92 & 16.91 & 27.16 \\
\cmark & \cmark & \cmark & \bf 88.17 & \bf 14.41 & \bf 40.21 & \bf 17.47 & \bf 27.49  \\
\bottomrule[1.2pt]
\end{tabular}
}
\label{tab:main-benchmark-stage}
\end{minipage}
\hspace{0.01\textwidth}
\begin{minipage}{0.3\textwidth}
\centering
\caption{\textbf{Effect of distillation techniques}. Here LD denotes logits distillation in Eq.~\ref{eq: logits distill} and AD denotes multi-head attention distillation in Eq.~\ref{eq: attn distill}.}
\resizebox{\linewidth}{!}{
\renewcommand\tabcolsep{15.0pt}
\begin{tabular}{ccc}
\toprule[1.2pt]
LD & AD & \bf MNLI \\
\midrule
\xmark & \xmark & 86.73 \\
\cmark & \xmark & 87.32 \\
\xmark & \cmark & 87.67 \\
\cmark & \cmark & \bf 88.17 \\
\bottomrule[1.2pt]
\end{tabular}
}
\label{tab:main-benchmark-distill}
\end{minipage}
\end{table*}

\section{Related Work}

\noindent\textbf{Quantization for LLMs}
Quantization~\citep{team2025minicpm4,du2024bitdistiller,ma2024bitnet158} has emerged as a widely adopted technique for enhancing the efficiency and scalability of LLMs. Post-training quantization (PTQ)~\citep{xiao2023smoothquant,dettmers2022int8} like GPTQ~\citep{frantar2022gptq} and AWQ~\citep{lin2024awq} has been extensively studied for weight-only quantization of LLMs. PTQ applies low-bit quantization to a full-precision model using a small set of calibration data, without requiring access to the end-to-end training loss. However, PTQ always suffer from significant performance degradation, especially when quantization bits are lower than 4 bits~\citep{dettmers2022int8}. To address this limitation, quantization-aware training (QAT)~\citep{team2025minicpm4,liu2023llm,chen2024efficientqat} has been introduced, which continues training the quantized LLMs with sufficient optimization, thereby raising the performance ceiling achievable by quantized models.

\noindent\textbf{Knowledge Distillation for LLMs}
Knowledge distillation~\citep{ko2024distillm,hinton2015distilling,wang2020minilm,team2025minicpm4} has proven to be an effective technique for compressing large language models (LLMs) while preserving accuracy, by transferring knowledge from a high-capacity teacher model to a more compact student model. More recently, it has also been shown effective for transferring knowledge from full-precision models to quantized LLMs. For example, TSLD~\citep{kim2023token} employs layer-to-layer distillation to enhance quantization-aware training (QAT) for ternary quantization, while BitDistiller~\citep{du2024bitdistiller} leverages self-distillation to improve the performance of LLMs at ultra-low precisions (e.g., 2 or 3 bits). Despite these advances, most existing methods primarily target general language modeling capabilities and still exhibit noticeable performance gaps in downstream applications compared to their full-precision counterparts.

\section{Conclusion}
In this work, we investigated the problem of adapting pre-trained LLMs to ultra-low precision with only 1.58-bit weights, motivated by the practical need to deploy large-scale models on edge devices under strict memory and latency constraints. To this end, we introduced \oursfull{}, a three-stage framework that first performs model refinement with SubLN, and then continued pre-training to recover critical representation capacity, followed by knowledge distillation at both the hidden-state and attention-relation levels to narrow the accuracy gap between low-precision students and high-precision teachers. Extensive experiments on multiple downstream tasks demonstrate that our method, \ours{}, achieves performance competitive with FP16 models while significantly reducing the computational and memory footprint. Beyond improving efficiency, our approach provides new insights into how low-bit quantization interacts with both pretraining and distillation dynamics, shedding light on scalable strategies for resource-constrained deployment.


\bibliographystyle{alpha}
\bibliography{main}

\newcommand{\etalchar}[1]{$^{#1}$}
\begin{thebibliography}{MWM{\etalchar{+}}24}

\bibitem[AAA{\etalchar{+}}23]{achiam2023gpt}
Josh Achiam, Steven Adler, Sandhini Agarwal, Lama Ahmad, Ilge Akkaya, Florencia~Leoni Aleman, Diogo Almeida, Janko Altenschmidt, Sam Altman, Shyamal Anadkat, et~al.
\newblock Gpt-4 technical report.
\newblock {\em arXiv preprint arXiv:2303.08774}, 2023.

\bibitem[BLC13]{bengio2013ste}
Yoshua Bengio, Nicholas L{\'e}onard, and Aaron Courville.
\newblock Estimating or propagating gradients through stochastic neurons for conditional computation.
\newblock {\em arXiv preprint arXiv:1308.3432}, 2013.

\bibitem[BMH{\etalchar{+}}22]{borgeaud2022improving}
Sebastian Borgeaud, Arthur Mensch, Jordan Hoffmann, Trevor Cai, Eliza Rutherford, Katie Millican, George van~den Driessche, Jean-Baptiste Lespiau, Bogdan Damoc, Aidan Clark, et~al.
\newblock Improving language models by retrieving from trillions of tokens, 2022.

\bibitem[CSX{\etalchar{+}}24]{chen2024efficientqat}
Mengzhao Chen, Wenqi Shao, Peng Xu, Jiahao Wang, Peng Gao, Kaipeng Zhang, and Ping Luo.
\newblock Efficientqat: Efficient quantization-aware training for large language models.
\newblock {\em arXiv preprint arXiv:2407.11062}, 2024.

\bibitem[DLBZ22]{dettmers2022int8}
Tim Dettmers, Mike Lewis, Younes Belkada, and Luke Zettlemoyer.
\newblock Gpt3. int8 (): 8-bit matrix multiplication for transformers at scale.
\newblock {\em Advances in neural information processing systems}, 35:30318--30332, 2022.

\bibitem[DLSZ21]{dettmers20218}
Tim Dettmers, Mike Lewis, Sam Shleifer, and Luke Zettlemoyer.
\newblock 8-bit optimizers via block-wise quantization.
\newblock {\em arXiv preprint arXiv:2110.02861}, 2021.

\bibitem[DZC{\etalchar{+}}24]{du2024bitdistiller}
Dayou Du, Yijia Zhang, Shijie Cao, Jiaqi Guo, Ting Cao, Xiaowen Chu, and Ningyi Xu.
\newblock Bitdistiller: Unleashing the potential of sub-4-bit llms via self-distillation.
\newblock {\em arXiv preprint arXiv:2402.10631}, 2024.

\bibitem[FAHA22]{frantar2022gptq}
Elias Frantar, Saleh Ashkboos, Torsten Hoefler, and Dan Alistarh.
\newblock Gptq: Accurate post-training quantization for generative pre-trained transformers.
\newblock {\em arXiv preprint arXiv:2210.17323}, 2022.

\bibitem[GYZ{\etalchar{+}}25]{guo2025deepseekr1}
Daya Guo, Dejian Yang, Haowei Zhang, Junxiao Song, Ruoyu Zhang, Runxin Xu, Qihao Zhu, Shirong Ma, Peiyi Wang, Xiao Bi, et~al.
\newblock Deepseek-r1: Incentivizing reasoning capability in llms via reinforcement learning.
\newblock {\em arXiv preprint arXiv:2501.12948}, 2025.

\bibitem[HKG{\etalchar{+}}15]{cnndm}
Karl~Moritz Hermann, Tomás Kociský, Edward Grefenstette, Lasse Espeholt, Will Kay, Mustafa Suleyman, and Phil Blunsom.
\newblock Teaching machines to read and comprehend.
\newblock In {\em NIPS}, pages 1693--1701, 2015.

\bibitem[HVD15]{hinton2015distilling}
Geoffrey Hinton, Oriol Vinyals, and Jeff Dean.
\newblock Distilling the knowledge in a neural network, 2015.

\bibitem[HZL{\etalchar{+}}24]{hou2024large}
Yupeng Hou, Junjie Zhang, Zihan Lin, Hongyu Lu, Ruobing Xie, Julian McAuley, and Wayne~Xin Zhao.
\newblock Large language models are zero-shot rankers for recommender systems.
\newblock In {\em European Conference on Information Retrieval}, pages 364--381. Springer, 2024.

\bibitem[KDSP25]{kostina2025large}
Arina Kostina, Marios~D Dikaiakos, Dimosthenis Stefanidis, and George Pallis.
\newblock Large language models for text classification: Case study and comprehensive review.
\newblock {\em arXiv preprint arXiv:2501.08457}, 2025.

\bibitem[KKCY24]{ko2024distillm}
Jongwoo Ko, Sungnyun Kim, Tianyi Chen, and Se-Young Yun.
\newblock Distillm: Towards streamlined distillation for large language models.
\newblock {\em arXiv preprint arXiv:2402.03898}, 2024.

\bibitem[KLL{\etalchar{+}}23]{kim2023token}
Minsoo Kim, Sihwa Lee, Janghwan Lee, Sukjin Hong, Du-Seong Chang, Wonyong Sung, and Jungwook Choi.
\newblock Token-scaled logit distillation for ternary weight generative language models.
\newblock {\em Advances in Neural Information Processing Systems}, 36:42097--42118, 2023.

\bibitem[Lin04]{lin2004rouge}
Chin-Yew Lin.
\newblock Rouge: A package for automatic evaluation of summaries.
\newblock In {\em Text summarization branches out}, pages 74--81, 2004.

\bibitem[LOZ{\etalchar{+}}23]{liu2023llm}
Zechun Liu, Barlas Oguz, Changsheng Zhao, Ernie Chang, Pierre Stock, Yashar Mehdad, Yangyang Shi, Raghuraman Krishnamoorthi, and Vikas Chandra.
\newblock Llm-qat: Data-free quantization aware training for large language models.
\newblock {\em arXiv preprint arXiv:2305.17888}, 2023.

\bibitem[LSK{\etalchar{+}}25]{lee2025unifying}
Jung~Hyun Lee, Seungjae Shin, Vinnam Kim, Jaeseong You, and An~Chen.
\newblock Unifying block-wise ptq and distillation-based qat for progressive quantization toward 2-bit instruction-tuned llms.
\newblock {\em arXiv preprint arXiv:2506.09104}, 2025.

\bibitem[LTT{\etalchar{+}}24]{lin2024awq}
Ji~Lin, Jiaming Tang, Haotian Tang, Shang Yang, Wei-Ming Chen, Wei-Chen Wang, Guangxuan Xiao, Xingyu Dang, Chuang Gan, and Song Han.
\newblock Awq: Activation-aware weight quantization for on-device llm compression and acceleration.
\newblock {\em Proceedings of machine learning and systems}, 6:87--100, 2024.

\bibitem[MWH{\etalchar{+}}25]{ma2025bitnet2b4t}
Shuming Ma, Hongyu Wang, Shaohan Huang, Xingxing Zhang, Ying Hu, Ting Song, Yan Xia, and Furu Wei.
\newblock Bitnet b1. 58 2b4t technical report.
\newblock {\em arXiv preprint arXiv:2504.12285}, 2025.

\bibitem[MWM{\etalchar{+}}24]{ma2024bitnet158}
Shuming Ma, Hongyu Wang, Lingxiao Ma, Lei Wang, Wenhui Wang, Shaohan Huang, Lifeng Dong, Ruiping Wang, Jilong Xue, and Furu Wei.
\newblock The era of 1-bit llms: All large language models are in 1.58 bits.
\newblock {\em arXiv preprint arXiv:2402.17764}, 1(4), 2024.

\bibitem[PMH{\etalchar{+}}23]{penedo2023refinedwebdatasetfalconllm}
Guilherme Penedo, Quentin Malartic, Daniel Hesslow, Ruxandra Cojocaru, Alessandro Cappelli, Hamza Alobeidli, Baptiste Pannier, Ebtesam Almazrouei, and Julien Launay.
\newblock The refinedweb dataset for falcon llm: Outperforming curated corpora with web data, and web data only, 2023.

\bibitem[PRWZ02]{papineni2002bleu}
Kishore Papineni, Salim Roukos, Todd Ward, and Wei-Jing Zhu.
\newblock Bleu: a method for automatic evaluation of machine translation.
\newblock In {\em Proceedings of the 40th annual meeting of the Association for Computational Linguistics}, pages 311--318, 2002.

\bibitem[QY{\etalchar{+}}25]{qwen2025qwen25}
Qwen, :, An~Yang, Baosong Yang, Beichen Zhang, Binyuan Hui, Bo~Zheng, Bowen Yu, Chengyuan Li, Dayiheng Liu, Fei Huang, Haoran Wei, Huan Lin, Jian Yang, Jianhong Tu, Jianwei Zhang, Jianxin Yang, Jiaxi Yang, Jingren Zhou, Junyang Lin, Kai Dang, Keming Lu, Keqin Bao, Kexin Yang, Le~Yu, Mei Li, Mingfeng Xue, Pei Zhang, Qin Zhu, Rui Men, Runji Lin, Tianhao Li, Tianyi Tang, Tingyu Xia, Xingzhang Ren, Xuancheng Ren, Yang Fan, Yang Su, Yichang Zhang, Yu~Wan, Yuqiong Liu, Zeyu Cui, Zhenru Zhang, and Zihan Qiu.
\newblock Qwen2.5 technical report, 2025.

\bibitem[RWX{\etalchar{+}}24]{ren2024representation}
Xubin Ren, Wei Wei, Lianghao Xia, Lixin Su, Suqi Cheng, Junfeng Wang, Dawei Yin, and Chao Huang.
\newblock Representation learning with large language models for recommendation.
\newblock In {\em Proceedings of the ACM web conference 2024}, pages 3464--3475, 2024.

\bibitem[RZLL16]{qnli}
Pranav Rajpurkar, Jian Zhang, Konstantin Lopyrev, and Percy Liang.
\newblock Squad: 100, 000+ questions for machine comprehension of text.
\newblock {\em CoRR}, abs/1606.05250, 2016.

\bibitem[SLL{\etalchar{+}}23]{sun2023text}
Xiaofei Sun, Xiaoya Li, Jiwei Li, Fei Wu, Shangwei Guo, Tianwei Zhang, and Guoyin Wang.
\newblock Text classification via large language models.
\newblock {\em arXiv preprint arXiv:2305.08377}, 2023.

\bibitem[SPW{\etalchar{+}}13]{sst2}
Richard Socher, Alex Perelygin, Jean Wu, Jason Chuang, Christopher~D. Manning, Andrew Ng, and Christopher Potts.
\newblock Recursive deep models for semantic compositionality over a sentiment treebank.
\newblock In {\em Proceedings of the 2013 Conference on Empirical Methods in Natural Language Processing}, pages 1631--1642, Seattle, Washington, USA, October 2013. Association for Computational Linguistics.

\bibitem[TKF{\etalchar{+}}25]{team2025gemma}
Gemma Team, Aishwarya Kamath, Johan Ferret, Shreya Pathak, Nino Vieillard, Ramona Merhej, Sarah Perrin, Tatiana Matejovicova, Alexandre Ram{\'e}, Morgane Rivi{\`e}re, et~al.
\newblock Gemma 3 technical report.
\newblock {\em arXiv preprint arXiv:2503.19786}, 2025.

\bibitem[TXL{\etalchar{+}}25]{team2025minicpm4}
MiniCPM Team, Chaojun Xiao, Yuxuan Li, Xu~Han, Yuzhuo Bai, Jie Cai, Haotian Chen, Wentong Chen, Xin Cong, Ganqu Cui, et~al.
\newblock Minicpm4: Ultra-efficient llms on end devices.
\newblock {\em arXiv preprint arXiv:2506.07900}, 2025.

\bibitem[WBH{\etalchar{+}}20]{wang2020minilmv2}
Wenhui Wang, Hangbo Bao, Shaohan Huang, Li~Dong, and Furu Wei.
\newblock Minilmv2: Multi-head self-attention relation distillation for compressing pretrained transformers.
\newblock {\em arXiv preprint arXiv:2012.15828}, 2020.

\bibitem[WMD{\etalchar{+}}23]{wang2023bitnet1bit}
Hongyu Wang, Shuming Ma, Li~Dong, Shaohan Huang, Huaijie Wang, Lingxiao Ma, Fan Yang, Ruiping Wang, Yi~Wu, and Furu Wei.
\newblock Bitnet: Scaling 1-bit transformers for large language models.
\newblock {\em arXiv preprint arXiv:2310.11453}, 2023.

\bibitem[WNB18]{mnli}
Adina Williams, Nikita Nangia, and Samuel Bowman.
\newblock A broad-coverage challenge corpus for sentence understanding through inference.
\newblock In {\em Proceedings of the 2018 Conference of the North American Chapter of the Association for Computational Linguistics: Human Language Technologies, Volume 1 (Long Papers)}, pages 1112--1122. Association for Computational Linguistics, 2018.

\bibitem[WSM{\etalchar{+}}19]{wang2018glue}
Alex Wang, Amanpreet Singh, Julian Michael, Felix Hill, Omer Levy, and Samuel~R. Bowman.
\newblock {GLUE}: A multi-task benchmark and analysis platform for natural language understanding.
\newblock In {\em International Conference on Learning Representations}, 2019.

\bibitem[WWD{\etalchar{+}}20]{wang2020minilm}
Wenhui Wang, Furu Wei, Li~Dong, Hangbo Bao, Nan Yang, and Ming Zhou.
\newblock Minilm: Deep self-attention distillation for task-agnostic compression of pre-trained transformers.
\newblock {\em Advances in neural information processing systems}, 33:5776--5788, 2020.

\bibitem[WZQ{\etalchar{+}}24]{wu2024survey}
Likang Wu, Zhi Zheng, Zhaopeng Qiu, Hao Wang, Hongchao Gu, Tingjia Shen, Chuan Qin, Chen Zhu, Hengshu Zhu, Qi~Liu, et~al.
\newblock A survey on large language models for recommendation.
\newblock {\em World Wide Web}, 27(5):60, 2024.

\bibitem[XLS{\etalchar{+}}23]{xiao2023smoothquant}
Guangxuan Xiao, Ji~Lin, Mickael Seznec, Hao Wu, Julien Demouth, and Song Han.
\newblock Smoothquant: Accurate and efficient post-training quantization for large language models.
\newblock In {\em International conference on machine learning}, pages 38087--38099. PMLR, 2023.

\bibitem[YLY{\etalchar{+}}25]{yang2025qwen3}
An~Yang, Anfeng Li, Baosong Yang, Beichen Zhang, Binyuan Hui, Bo~Zheng, Bowen Yu, Chang Gao, Chengen Huang, Chenxu Lv, et~al.
\newblock Qwen3 technical report.
\newblock {\em arXiv preprint arXiv:2505.09388}, 2025.

\bibitem[ZSB{\etalchar{+}}24]{zhao2024optimizing}
Yiyun Zhao, Prateek Singh, Hanoz Bhathena, Bernardo Ramos, Aviral Joshi, Swaroop Gadiyaram, and Saket Sharma.
\newblock Optimizing llm based retrieval augmented generation pipelines in the financial domain.
\newblock In {\em Proceedings of the 2024 Conference of the North American Chapter of the Association for Computational Linguistics: Human Language Technologies (Volume 6: Industry Track)}, pages 279--294, 2024.

\end{thebibliography}




\newpage

\end{document}